\def\year{2018}\relax
\documentclass[letterpaper]{article} 
\usepackage{aaai18}  
\usepackage{times}  
\usepackage{helvet}  
\usepackage{courier}  
\usepackage{url}  
\usepackage{graphicx}  
\newcommand*\rot{\rotatebox{90}}
\usepackage{mathtools}
\usepackage{amsthm}
\usepackage[normalem]{ulem}
\usepackage{csquotes}
\usepackage{tabularx}
\usepackage{multicol}
\usepackage{amsfonts}
\usepackage{enumitem}

\begin{document}
%
\title{A Deep Cascade Network for Unaligned Face Attribute Classification}
\author{Hui Ding,$^1$ Hao Zhou,$^2$ Shaohua Kevin Zhou,$^3$ Rama Chellappa$^4$\\
$^{1,2,4}$University of Maryland, College Park\\
$^3$Siemens Healthineers, New Jersey\\
}
\maketitle
\begin{abstract}
Humans focus attention on different face regions when recognizing face attributes. Most existing face attribute classification methods use the whole image as input. Moreover, some of these methods rely on fiducial landmarks to provide defined face parts. In this paper, we propose a cascade network that simultaneously learns to localize face regions specific to attributes and performs attribute classification without alignment. First, a weakly-supervised face region localization network is designed to automatically detect regions (or parts) specific to attributes. Then multiple part-based networks and a whole-image-based network are separately constructed and combined together by the region switch layer and attribute relation layer for final attribute classification. A multi-net learning method and hint-based model compression is further proposed to get an effective localization model and a compact classification model, respectively. Our approach achieves significantly better performance than state-of-the-art methods on unaligned CelebA dataset, reducing the classification error by 30.9\%.

\end{abstract}

\section{Introduction}
Face attributes describe the characteristics observed from a face image.
They were first introduced by \citeauthor{kumar2009attribute}~\shortcite{kumar2009attribute} as mid-level features for face verification~\cite{kumar2011describable} and since then have attracted much attention. 
The last few years have witnessed their successful applications in hashing~\cite{li2015two}, face retrieval~\cite{siddiquie2011image}, and one-shot face recognition~\cite{jadhav2016deep}. Recently, researchers have begun to investigate the possibility of synthesizing face images based on face attributes~\cite{radford2015unsupervised,yan2016attribute2image}.

Despite their wide applications, face attribute recognition is not an easy task. One reason is that recognizing different face attributes may require attentions to different regions of the face~\cite{moran1985selective,posner1990attention}. For example, local attributes like \textit{Mustache} could be recognized by just checking the region containing the mouth. Remaining face region does not provide useful information and may even hamper this particular attribute recognition.
However, recognizing global attributes like \textit{Pale Skin} may require information from the whole face region. Most current research studies do not pay special attention to this problem. They either detect facial landmarks and extract hand-crafted features from patches around them~\cite{kumar2009attribute,berg2013poof} or train a deep network to classify the attributes by taking a whole face as input~\cite{liu2015deep,wang2016walk,rudd2016moon,hand2016attributes}. 


In this paper, we propose a learning-based method that dynamically selects different face regions for unaligned face attribute prediction.
It integrates two networks using a cascade: a face region localization (FRL) network followed by an attribute classification network. The localization network detects face areas specific to attributes, especially those that have local spatial support. The classification network selectively leverages information from these face regions to make the final prediction.

For accurate face region detection, our localization network is constructed under a multi-task learning framework. The lower layers which are used to extract low level features are shared by all the tasks while the high-level semantics are learned separately. Moreover, a global average pooling is applied to force the network to learn location-sensitive information~\cite{lin2013network}. Although the network is trained in a weakly-supervised manner with attribute labels only, the detected face regions are consistent with what one may expect.
As a result, face alignment algorithms which are usually sensitive to occlusion, variations of pose and illumination are not needed. 

For each face region (also called a part) detected by our localization network, we train a separate attribute classification network, called a part-based subnet. 
The localized face parts may not contain enough contextual information for predicting global attributes.
Thus, a whole-image-based subnet is also trained. To combine the information from the part-based and whole-image-based subnets, a two-layer fully-connected classifier is built on top of the output attribute scores. The first layer is used to select the relevant subnet for predicting each attribute, while the second layer is designed to model the rich attribute relations. The integrated system is called the parts and whole (PaW) network.

Since the face region localization network is supervised by attribute labels, it is appealing to adapt its weights to initialize the subnets in PaW. However, features from the localization network, which are mainly designed for localization purpose, are generally not very discriminative for attribute classification. To this end, a multi-net learning method is proposed. It utilizes a network with enhanced attribute classification capability to train the localization network to find a more discriminative solution. 

A naive implementation of the PaW network is problematic since the number of total parameters increases linearly with the number of attributes, and the subnet adapted from the FRL network is not very compact. To jointly train the PaW network end-to-end, a hint-based model compression technique is further proposed. This not only leads to a compact model with only $11M$ parameters, but also reduces the training time significantly.

We applied the proposed method to CelebA dataset~\cite{liu2015deep}. With no use of alignment information, our method achieves an accuracy of \textbf{91.23}\%, reducing the classification error by a significant margin of \textbf{30.9}\% compared with state-of-the-art~\cite{liu2015deep}. 
Moreover, our designed model could select the most relevant face region for predicting each face attribute.

To summarize, the contributions of this paper are listed below: 

\begin{itemize}[noitemsep]
  \item A weakly-supervised localization network is designed to accurately locate attribute regions.
  \item A hybrid classification network is proposed to dynamically choose the pertinent face regions for predicting different attributes.
  \item A hint-based model compression technique is explored to obtain a compact model.
  \item The state-of-the-art of unaligned face attribute classification is significantly improved by the proposed method.
\end{itemize}

\section{Related Works}
\begin{figure*}
\centering
\includegraphics[width=\textwidth]{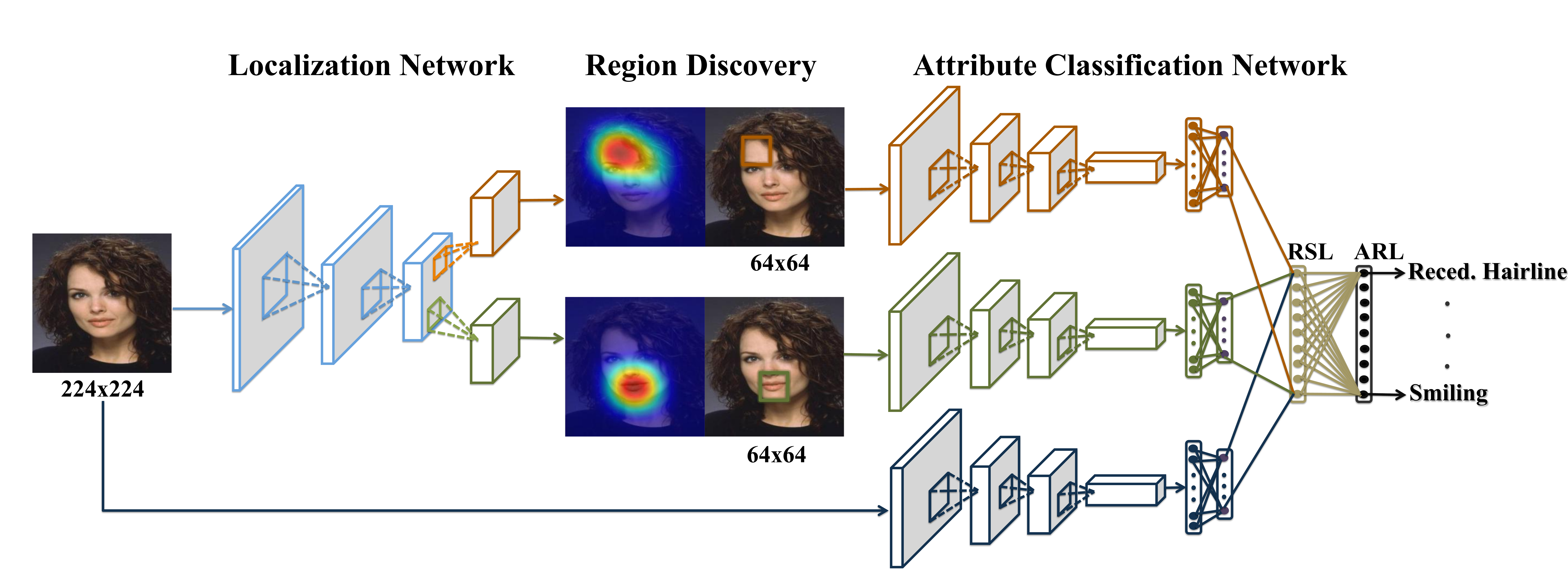}
\caption{Overview of our face attribute recognition framework. It consists of
a facial region localization (FRL) network and a Parts and Whole (PaW) classification network.  The localization network
detects a discriminative part for each attribute. Then the detected face regions and the whole face image are fed into 
the PaW classification network. The region switch layer (RSL) selects the relevant subnet for predicting the attribute, while the attribute relation layer (ARL) models the attribute relationships.
}
\label{fig:pipeline}
\end{figure*}

\noindent \textbf{Face Attribute Recognition}
Early works~\cite{kumar2009attribute,berg2013poof} on face attribute recognition used manually defined face parts to extract features and then train a linear SVM classifier. This strategy though is well suited for near-frontal faces, is heavily dependent on the accuracy of landmark detection. 
Recently, with the emergence of large-scale data and deep neural networks, 
holistic methods~\cite{liu2015deep,wang2016walk,huang2016learning} have produced better performance than the part-based method. 
~\citeauthor{liu2015deep}\shortcite{liu2015deep} noticed that a deep model pre-trained for face recognition implicitly learns attributes. 
\citeauthor{huang2016learning}~\shortcite{huang2016learning} employed a quintuplet loss to combat the imbalanced data distribution problem. 
These methods typically use the whole face image to train a deep network, ignoring the fact that different facial attributes have different attentional facial regions. This problem has been recently noticed 
in~\cite{ehrlich2016facial,murrugarralearning}.
\citeauthor{murrugarralearning}~\shortcite{murrugarralearning} created human gaze maps for each attribute such that only features within the saliency maps are used for attribute recognition. Our method differs from the aforementioned approaches in the sense that \textit{the face parts are localized automatically without relying on detected landmarks or human gaze data.} Moreover, our classification network can dynamically select the attentional face regions for predicting different attributes.

\noindent\textbf{Weakly Supervised Object Localization} 
Despite training with only image-level labels, recent 
works~\cite{oquab2015object,zhou2016learning,cinbis2017weakly}
showed that deep Convolutional Neural Networks (CNN) have remarkable object localization ability. 
\citeauthor{zhou2016learning}~\shortcite{zhou2016learning} proposed a class activation mapping method to localize the objects with class labels only. The design of our face region localization network is motivated by this work. 
However, to fully utilize the correlations among different face attributes, the localization network is designed in a multi-task learning framework. 

\noindent \textbf{Model Compression}
To obtain a compact model, several methods including network distillation~\cite{buciluǎ2006model}, parameter pruning~\cite{lecun1989optimal} have been proposed. 
Recently, knowledge distillation~\cite{hinton2015distilling} has been shown to be very effective to teach a small student model. 
However, it can not be directly applied to our problem: the teacher net uses the soft labels which contain rich ambiguous information to supervise the student net, while for attribute classification, the output has only one logit for each attribute. Thus, a new loss function based on hints is proposed to replace soft label supervision.

\section{Proposed Method}
The proposed method contains two networks: a localization network and an attribute classification network. An overview of the framework is shown in Figure~\ref{fig:pipeline}. 
First, we adopt the multi-net learning method to train a face region localization (FRL) network. 
Then one attentional region is detected for each attribute by the FRL network, which is fed into the PaW network for attribute prediction. To train the PaW end-to-end, a hint-based method is further applied to compress the model. 
The details of the proposed approach are discussed below.

\subsection{Face Region Localization (FRL) Network}
One challenge in designing a face region localization algorithm is that we do not have the labeled regions available. \citeauthor{murrugarralearning}~\shortcite{murrugarralearning} used human gaze to label the related region for each attribute, however, this is both time consuming and expensive. Inspired by the success in weakly supervised object localization~\cite{zhou2016learning}, we apply a global average pooling (GAP) network for the localization task, and train it in a weakly-supervised way where only face attribute labels are needed. In this network structure, a GAP layer is used to pool features from the last convolutional layer, and a fully-connected layer is followed to predict the attribute score. A localization heatmap, $H_j$, for the $j$-th attribute, is obtained by applying the class activation mapping method. $H_j = \sum_{i=1}^{N}w_{j,i} F_i, i=1,...,N$, where $F_i$ is the output feature maps from the last convolutional layer and $w_{j,i}$ is the $i$-th weight of the fully connected layer for predicting the $j$-th attribute. $N$ is chosen to be $32$ in our experiments.

We design the FRL network using multi-task learning~\cite{caruana1998multitask} strategy, where each attribute can be seen as one separate task. It has five VGGNet~\cite{simonyan2014very} convolutional modules shared by all the attributes, and a domain adapted convolutional layer which has $M$ different branches for each attribute, where $M=40$ is the number of face attributes.  The weights of the network are initialized from the VGG-Face CNN~\cite{parkhi2015deep} which is trained on a large-scale face recognition dataset.

\subsubsection{Multi-Net Learning} 
Since the supervision of the FRL network comes from the attribute tags, it is appealing to transfer its weights to the subnets in PaW for faster convergence and better performance. However, training the FRL net in a plain way leads to less discriminative features due to GAP regularization~\cite{zhou2016learning}. This is also verified in our experiments.
To this end, a multi-net learning (MNL) method is proposed to boost the classification performance of the GAP feature, which yield improved final attribute classification.

\begin{figure}[!tb]
\centering
\includegraphics[width=0.45\textwidth]{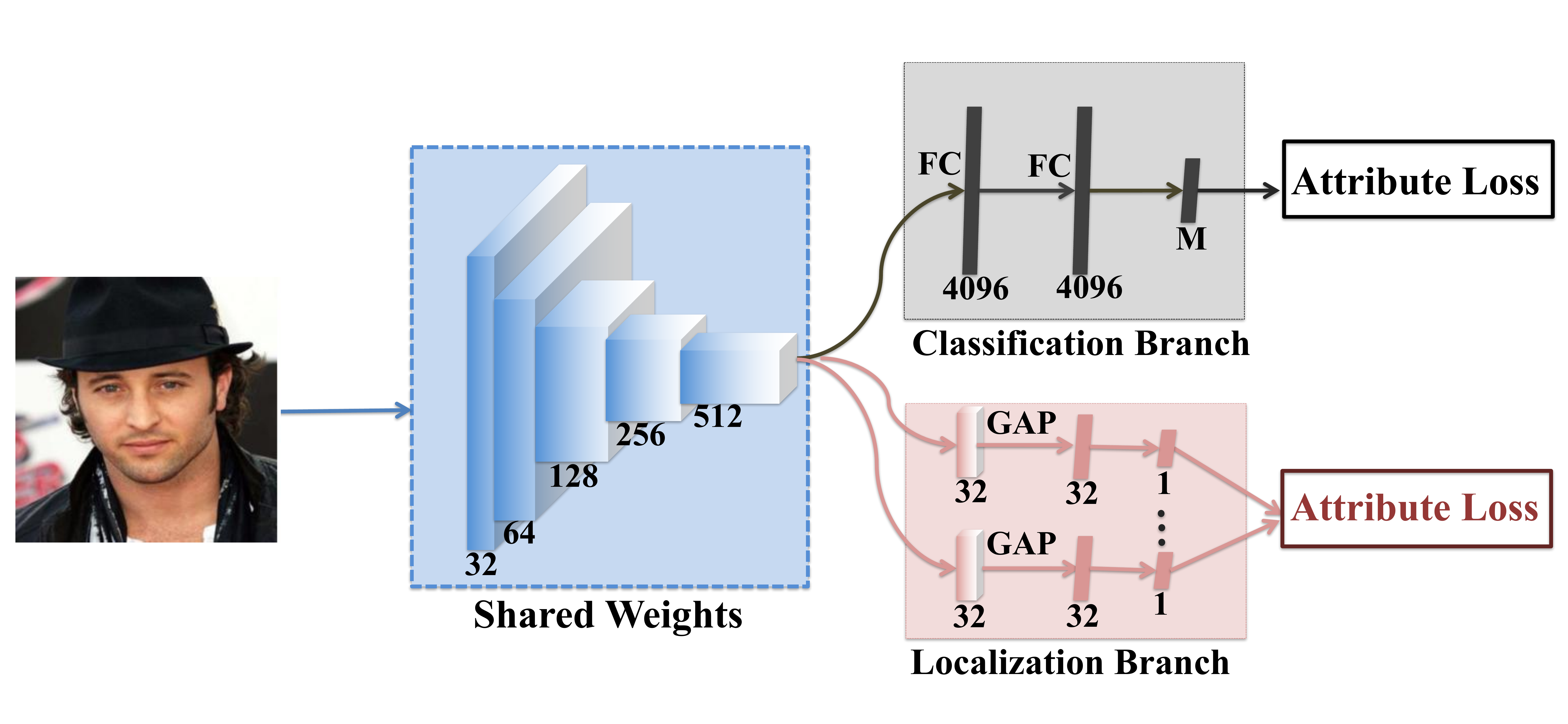}
\caption[face CNN resp]{Multi-Net Learning.}
\label{fig:mnl}
\end{figure}

The network architecture for MNL is shown in Figure~\ref{fig:mnl}. 
Except for the FRL network (blue and red boxes), another two fully-connected layers (gray box) are also attached to the output of the fifth convolutional module. 
We call it a classification branch because of its improved performance on the classification task compared with the localization branch.
The idea is to simultaneously train the two different types of networks with the same attributes loss. Meanwhile the first several convolutional layers are constructed to be shared between them. The gradients from both classification and localization branches are backpropagated to the shared layers. This extra supervision from the classification branch regularizes the training process to search for a more discriminative solution.
Interestingly, we find this simple learning strategy is beneficial for both branches in terms of classification performance. 
\textit{After the multi-net training is completed, the classification branch is removed, and only the localization branch is kept for extracting attribute-specific heatmaps.}

To localize the face region, we upsample the location heatmap to the original image size $224\times 224$, and find the position that corresponds to the maximum value. Then, a $64\times64$ patch centered around this position is cropped from the original image as the detected face region. We empirically found this patch size to be sufficient for most face parts. This process is repeated for each attribute and $M$ face regions are obtained.

\subsection{Attribute Classification Network} 
As shown in Figure~\ref{fig:pipeline}, the proposed attribute classification network PaW contains $M$ part-based subnets and one whole-image-based subnet. After getting the predicted attributes scores from each subnet, a two-layer fully-connected classifier is adopted to combine them.

\subsubsection{Parts and Whole (PaW) Classification Network}
Suppose $x_0$ represents the whole face image, $x_1,..., x_M$ represent face region related to each face attribute. $g_i, i\in{0,.., M}$ represent the ($M+1$) subnets. Each $x_i$ is first fed into its corresponding subnet $g_i$ to predict the $M$ attribute scores $\{s_{i,j}\}$, where $s_{i,j}$ represents the predicted score of the $j$-th attribute by the $i$-th subnet. The reason why we train each part-based subnet to predict $M$ attributes instead of the one related to the input region is based on the observation that some attributes can usually be predicted by other ones~\cite{torfasonface}. The predicted scores $s_{i,j}$ will be fed into a region switch layer (RSL) which is designed as $r_j=\sum_{i=0}^{M} W_{ij}s_{ij}, j=1,...,M, W\in R^{(M+1)\times M}$ whose element in the $i$-th row and $j$-th column is $W_{ij}$. RSL adopts a group fully-connected structure, where the $j$-th output is only connected with the $j$-th attribute scores predicted by all subnets.
Especially, it could balance the scores from the part-based and whole-image-based subnets by putting more weight to the one that is more important. An attribute relation layer (ARL), which is a fully-connected layer, then takes these $r_j, j\in{1,..., M}$ as input to predict the final score for each face attribute. ARL here is used to further model the high correlations among the face attributes.
The PaW network is trained end-to-end with the sigmoid cross entropy loss: $L_{attr} = \sum_{j=1}^M y_{j}\log o_{j} + (1-y_{j})\log (1-o_{j})$, where $y_j$'s are the attributes labels, and $o_j$'s are the outputs from the ARL layer.

\subsubsection{Hint-based Model Compression}
\begin{figure}[!tb]
\centering
\includegraphics[width=0.35\textwidth]{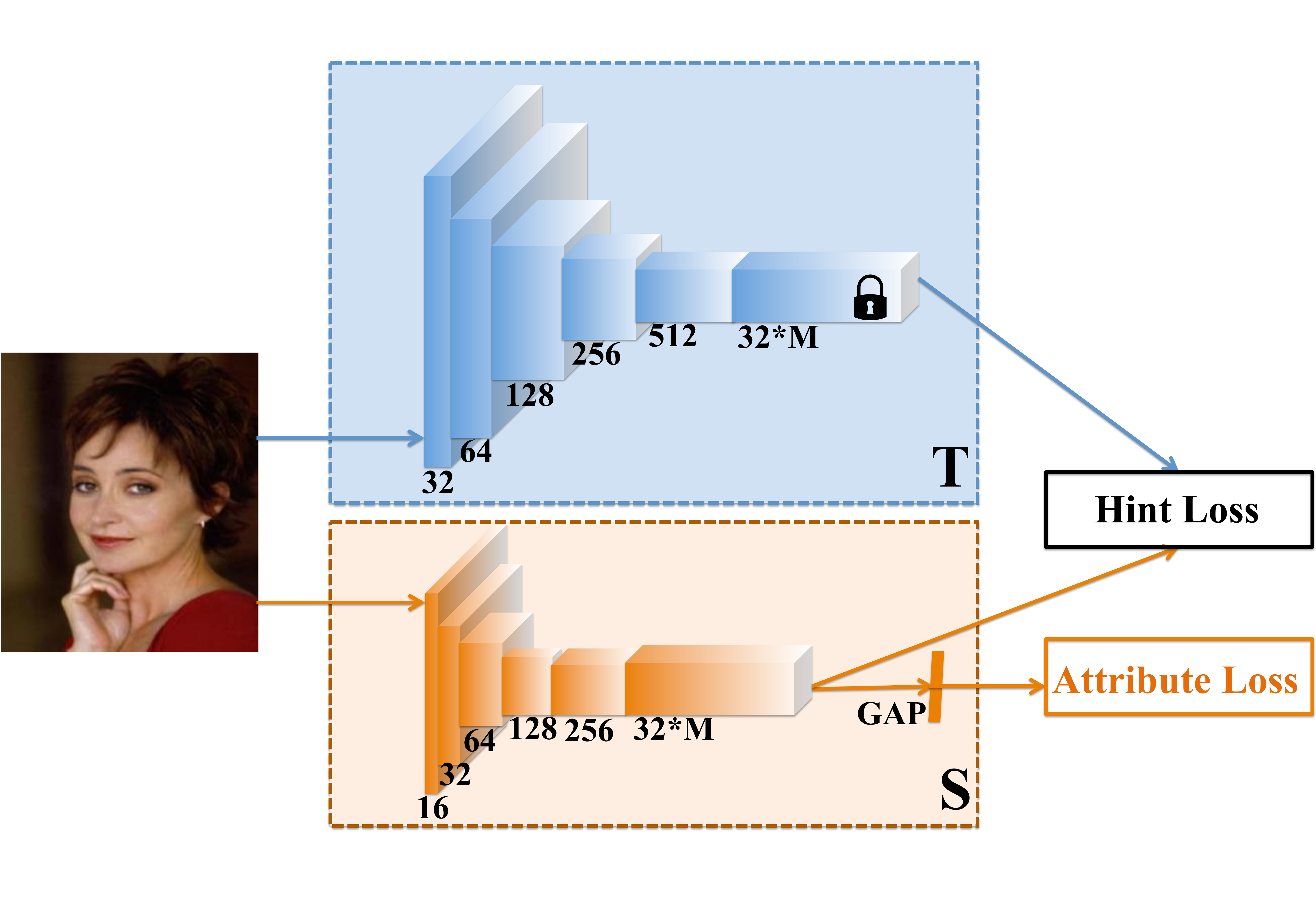}
\caption[face CNN resp]{Hint-based Model Compression.}
\label{fig:compress}
\end{figure}

Training the PaW network in a naive way is both memory demanding and time consuming, since the total number of network parameters increases substantially as the number of attributes becomes large, and the subnet architecture adapted from the FRL network is not very compact.
To obtain a compact subnet model, we further propose a model compression technique. 
Motivated by~\cite{abu1992method,ding2016facenet2expnet}, we design a hint loss to make the student net (SNet) reconstruct the feature maps from the teacher net (TNet). It can be expressed as:
\begin{equation}
L_{hint}(w) = ||T_k(I) - S_l(I,w)||_2,
\end{equation}
where $k$ ($l$) is the chosen layer of the teacher (student) net to transfer (add) supervision, $w$ are the weights of the student net to be learned, and $I$ is the input whole face image.
The network architecture is shown in Figure~\ref{fig:compress}. Besides the hint loss, the student network is also supervised by the attributes loss. Thus, the total loss function can be written as $L_S = \lambda_1 L_{hint} + \lambda_2 L_{attr}$. 
The FRL network trained by MNL is adopted as the teacher network to teach the whole-image-based subnet (or the student net). Since it is fully-convolutional and deeper layer generally captures high-level semantics~\cite{zeiler2014visualizing,escorcia2015relationship}, we set the supervision layer $k$ to be the teacher network's last convolutional layer. During training, the weights of the teacher network are frozen, and only the student network is learned. The whole training is carried out in two stages: first setting $\lambda_1=1,\lambda_2=0$, and training S with only the hint loss. In this way, the knowledge of the teacher network could help the student network find a good initialization. Then we set $\lambda_1=0,\lambda_2=1$ and train S with attribute loss only. After the whole-image-based subnet is learned, its weights are used to initialize all the part-based subnets in PaW.

\subsection{Training Methodology}
The whole training process is carried out as follows: 
\begin{enumerate}[noitemsep]
\item First, MNL is adopted to train the FRL network with superior classification performance;
\item Then hint-based compression method is applied to train a compact whole-image-based subnet $g_0$ using the learned FRL network as the teacher net.
\item Initialize each part-based subnet $\{g_i\}_{i=1}^M$ using the weights from $g_0$ and then train each subnet $g_i$ independently using the corresponding attentional face region; 
\item By fixing all the part-based subnets and the whole-image-based subnet, the RSL and ARL are learned;
\item Finally, the PaW network is fine-tuned by back-propagating errors from ARL to all the lower layers of the part-based subnets and the whole-image-based subnet.
\end{enumerate}
All the subnets and the two layer fully-connected model are trained under the supervision of attribute labels. The third and forth steps initialize the classification model to be close to a good local minimum, which is important for the successful training of PaW.

\section{Experiments}
\subsection{Dataset}
We use the CelebA dataset~\cite{liu2015deep} in our experiments, since it has been widely used for face attributes classification. It consists of 202,599 face images collected from the Internet and annotated with 40 binary attributes. As suggested in~\cite{liu2015deep},  162,770 of these images are used for training, 19,867 and 19,962 are reserved for validation and testing respectively. 
Both unaligned and aligned sets are provided and we applied our method on the unaligned one (\textbf{uCelebA}).
To conduct experiments on uCelebA, we use the publicly available face detector~\cite{zhang2016joint} to detect faces. For 560 images which have no face detected, we use the provided landmarks to get the groundtruth bounding box (we empirically expand the minimum bounding box containing all landmarks twice to cover the neck and hair region). For 15,181 images with multiple faces detected, we select the bounding box that has maximum overlap with the groundtruth bounding box. This is the only preprocessing step applied to the unaligned images.

\subsection{Implementation details}
We applied MNL to train the FRL network. The learning rate is fixed to be 0.0001, and the network is trained for 10 epochs with batch size of 128. The FRL network is then compressed with a learning rate of $1e^{-7}$ for the hint loss training and 0.0001 for the attribute loss training.
The part-based subnets are trained for 15 epochs with the weights initialized from the whole-image-based subnet. After that, the RSL and ARL are trained with a learning rate of 0.1 with all subnets fixed. Finally, a learning rate of 0.001 is applied to train the PaW network in an end-to-end manner. Stochastic gradient descent (SGD) is used to train all the networks. The momentum and weight decay are set at $0.9$ and $0.0005$ for all the experiments respectively. Horizontal flipping is applied for data augmentation. We use Caffe~\cite{jia2014caffe} to implement our networks. 

\subsection{Ablative Analysis}
\subsubsection{Face Region Localization}
\begin{figure}[!ht]
   \centering
     \includegraphics*[width=\linewidth]{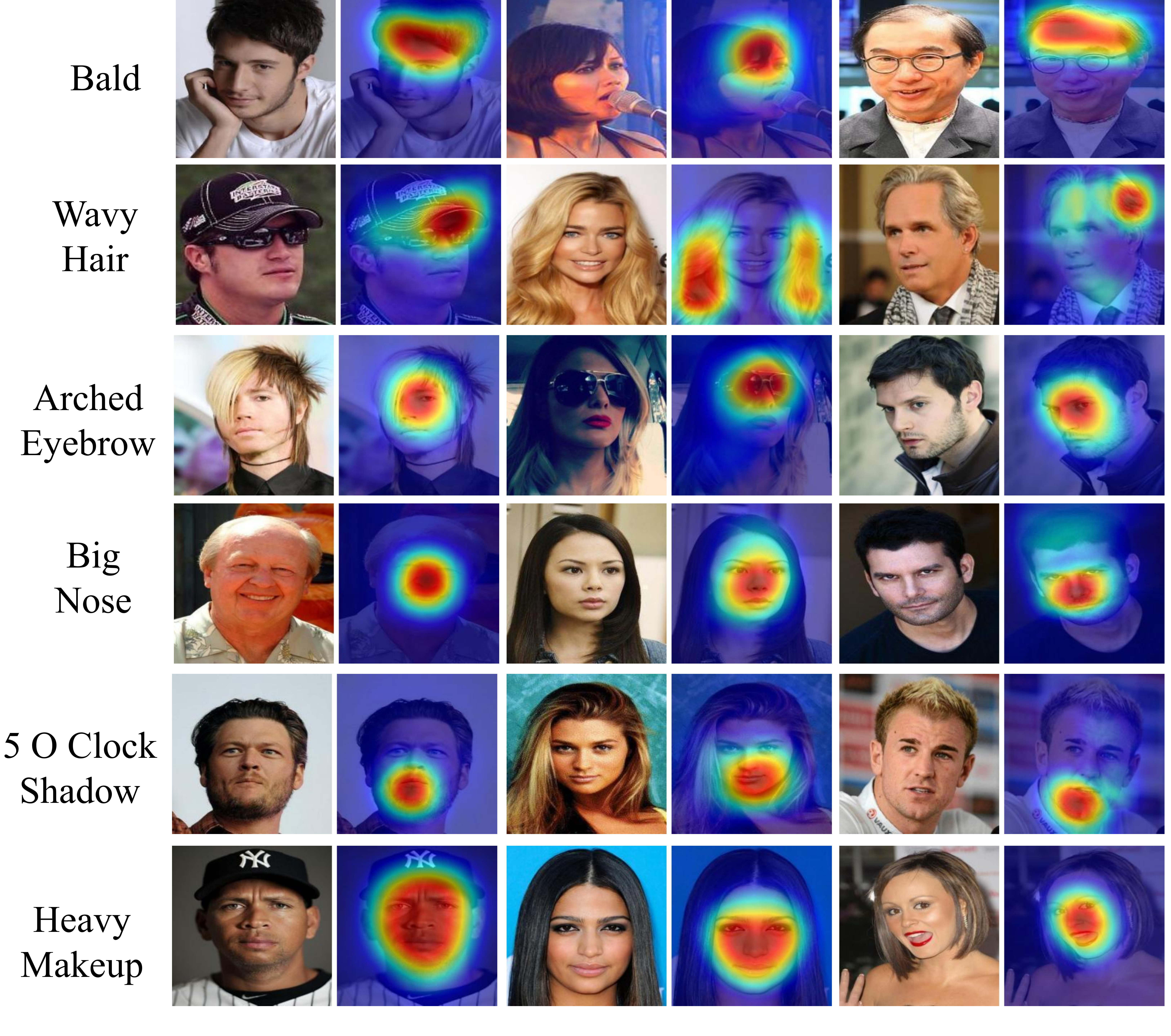}
\caption{Location heatmaps from the face region localization network. Face regions that correlate with facial attributes are discovered. }
\label{fig:big_pic}
\end{figure}

In this section, we evaluate the FRL network qualitatively. Figure~\ref{fig:big_pic} shows the location heatmaps corresponding to several attributes.
We observe that the localized parts are quite semantically meaningful, even though some face images have large pose variations or under occlusion. For example, the eye area produces the highest response for the \textit{Arched Eyebrow} attribute even though the woman wears sunglasses. While for the attribute of \textit{Wavy Hair}, the network localizes the head region although the man wears a hat. We also examine it quantitatively in the \textbf{Classification Results} section to show that accurate region localization is essential for good classification results. 

\subsubsection{Multi-Net Learning}
\begin{table}
\caption{Average classification accuracy on uCelebA dataset.}
\label{tab:acc}
\centering
\begin{tabular}{c|c|c}
\hline
\hline
Methods &Classif. Branch & Loc. Branch \\
\hline
Without MNL&-&91.01\\
MNL& 91.05&\textbf{91.07}\\
\hline
\end{tabular}
\end{table}

In this section, we study the ability of MNL for obtaining a localizable and discriminative deep representation. 
Table~\ref{tab:acc} summarizes the attribute classification results from classification and localization branches. 
We find that MNL consistently improves the classification performance of the localization branch, achieving an accuracy of $91.07\%$ vs. $91.01\%$ with/without MNL. 

\begin{table}
\caption{Fine-grained classification accuracy on CUB-200 dataset.}
\label{tab:finegrained}
\centering
\setlength\tabcolsep{2pt}
\begin{tabular}{c|c|c}
\hline
\hline
Methods &Classif. Branch & Loc. Branch \\
\hline
Without MNL on full image   & -&67.40\\
MNL on full image &72.10&\textbf{71.66}\\
\hline
Without MNL on crop   &-&71.90\\
MNL on crop &75.76&\textbf{76.03}\\
\hline
\end{tabular}
\end{table}
To further test the proposed MNL, we applied it on the popular CUB-200-2011 dataset~\cite{wah2011caltech} for fine-grained object recognition. The dataset contains 11,788 images, with 5,994 images for training and 5,794 for testing. The network architecture is the same as the one used in uCelebA, except that the last layer is replaced with 200 output nodes (the number of classes). 
The weights are initialized from VGGNet~\cite{simonyan2014very}. Table~\ref{tab:finegrained} summarizes the results. 
We find that the localization branch performs worse than the classification branch, with almost $4\%$ performance gap. After applying MNL, the accuracy of the localization branch is improved from $67.40\%$ to $71.66\%$ when using the full image. We also adopt the same localization technique as~\cite{zhou2016learning} to identify the bounding box of the birds in both the training and testing sets. With the cropped bird images as training data, the performance of the localization branch is further improved from $71.90\%$ to $76.03\%$. This further demonstrates that MNL is able to improve the discriminativeness of the GAP-based localization network.

\begin{figure*}
\centering
\includegraphics[width=\textwidth]{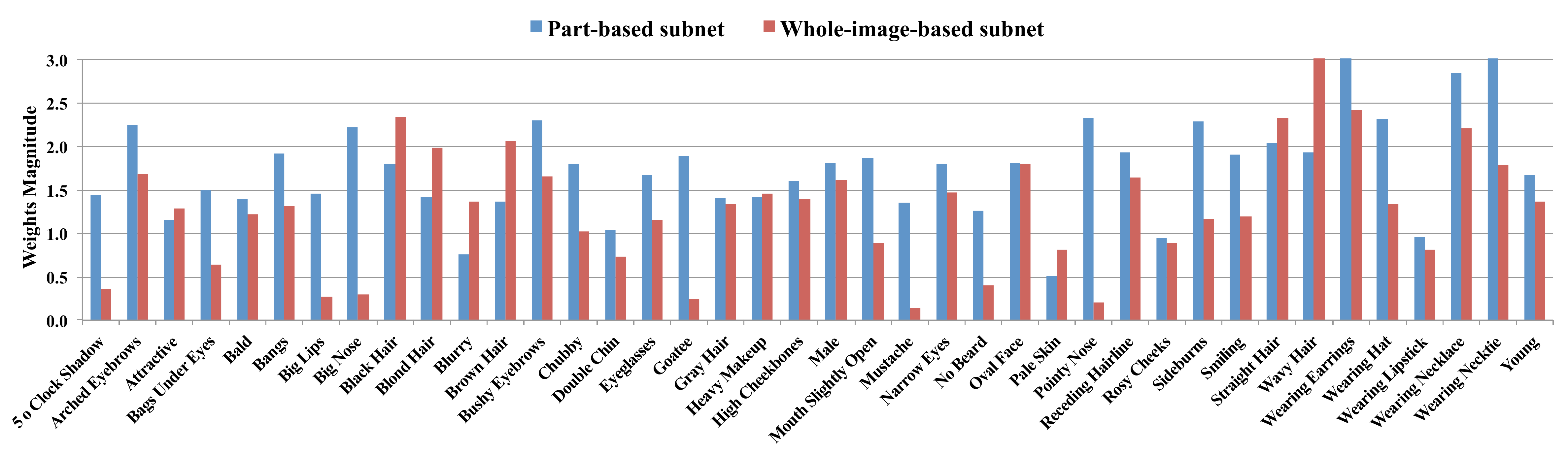}
\caption{Visualization of the region switch layer weights. For each attribute, the blue and the red bar represent the weight values of RSL that corresponds to the part-based subnet and whole-image-based subnet respectively. It shows that the weights of the part-based subnets are higher for the local attributes. For global attributes, the whole-image-based subnet is assigned larger weight.}
\label{fig:comb_w}
\end{figure*}
\subsubsection{Hint-based Model Compression}
\begin{table}
\caption{Comparison of average accuracy and compactness between different compressed models on uCelebA dataset.}
\label{tab:architect}
\centering
\setlength\tabcolsep{1pt}
\begin{tabular}{l|l|l|l|l}
\hline
\hline
Layer & TNet & SNet1 & SNet2 & SNet3 \\
\hline
Conv1& 3x3x32(2) &3x3x32 & 3x3x32 &3x3x16 \\
Pool1& 2x2x32 & 2x2x32 & 2x2x32 &2x2x16 \\
Conv2& 3x3x64(2) & 3x3x64 & 3x3x64 &3x3x32 \\
Pool2& 2x2x64 & 2x2x64 & 2x2x64 &2x2x32 \\
Conv3& 3x3x128(3) & 3x3x128 &3x3x128 &3x3x64  \\
Pool3& 2x2x128 & 2x2x128 &2x2x128 &2x2x64 \\
Conv4& 3x3x256(3) & 3x3x256 &3x3x256 &3x3x128 \\
Pool4& 2x2x256 & 2x2x256 &2x2x256 &2x2x128 \\
Conv5& 3x3x512(3) & 3x3x512 &3x3x512 &1x1x1280 \\
Conv6& 3x3x1280 & 3x3x1280 &1x1x1280 & n/a\\
Classifier&GAP&GAP&GAP&GAP\\
&FC40&FC40&FC40&FC40\\
\hline
Accuracy & 91.07& 91.02&90.89 &90.60\\
\hline
Param. & 19M & 6M& 2M& 0.27M \\
\hline
\end{tabular}
\end{table}
In this section, we analyze the effectiveness of our model compression technique. To show the flexibility and robustness of our method, we experiment with three student nets (SNet1, SNet2 and SNet3) with different sizes. 
Table~\ref{tab:architect} shows the network architectures and their classification results. We use $s\times s\times n(t)$ to denote kernel size $s\times s$ with $n$ output feature maps, where $t$ is the number of repeated convolution modules. We observe that the proposed method is able to compress a deep network to a relatively shallow network, with little performance drop. For SNet3, which achieves an accuracy of 90.60\%, the depth is shortened from 14 to 5, and the number of parameters is reduced from 19M to 0.27M. 


To further compare our approach with existing methods, we also train our models on the \textit{aligned} CelebA dataset. The results are summarized in Table~\ref{tab:compress}. We find that our SNet3 model achieves similar or better accuracy compared to these state-of-the-art methods, while being much more compact and thus faster.

\begin{table}
\caption{Comparison of average accuracy and compactness on the aligned CelebA dataset.}
\label{tab:compress}
\centering
\begin{tabular}{c|c|c}
\hline
\hline
Method & Accuracy & Param.\\
\hline
SOMP~\cite{lu2016fully}-thin-32&89.96&0.22M\\
SOMP~\cite{lu2016fully}-branch-32&90.74&1.49M \\
Low Rank~\cite{denton2014exploiting} &90.88 & 4.52M\\
\hline
SNet3&\textbf{90.89}&\textbf{0.27M}\\
\hline
\end{tabular}
\end{table}

\begin{figure}[!ht]
   \centering
     \includegraphics*[width=\linewidth]{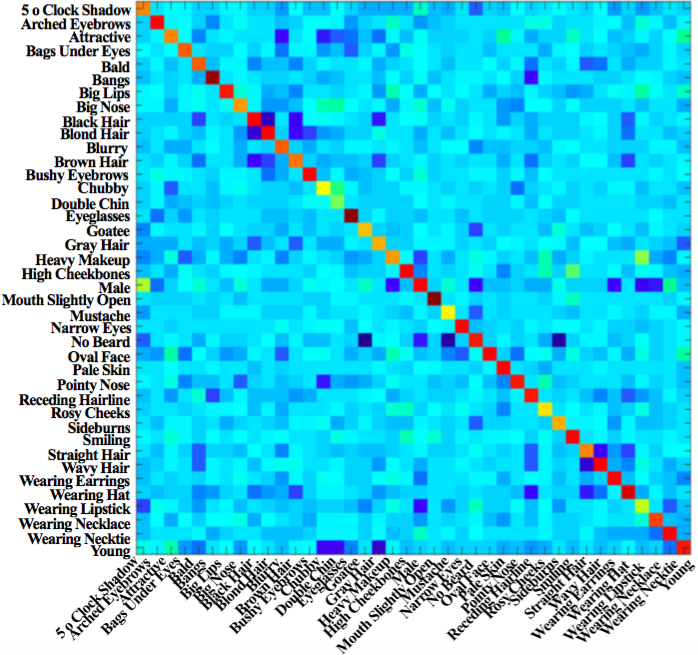}
   \caption{Attribute relation weights learned on uCelebA dataset. Red and yellow colors indicate high values while blue and green colors denote low values.}
   \label{fig:attr_corr}
\end{figure}

\noindent\begin{table*}[t]
\caption{Performance comparison with state of the art methods on 40 binary facial attributes. The best results are shown in bold.}
\resizebox{\textwidth}{!}{\begin{tabular}{c|c |c |c| c| c| c| c| c |c |c| c| c| c| c| c |c |c| c| c| c| c|c|}
\hline\hline
 &&\rot{5 o Clock Shadow} &\rot{Arched Eyebrows} &\rot{Attractive} &\rot{Bags Under Eyes} &\rot{Bald} &\rot{Bangs} &\rot{Big Lips} &\rot{Big Nose} &\rot{Black Hair} &\rot{Blond Hair} &\rot{Blurry} &\rot{Brown Hair} &\rot{Bushy Eyebrows} &\rot{Chubby} &\rot{Double Chin} &\rot{Eyeglasses} &\rot{Goatee} &\rot{Gray Hair} &\rot{Heavy Makeup} &\rot{High Cheekbones} &\rot{Male} \\
 \hline 
 
 & LNets+ANet~\cite{liu2015deep} & 91.00 & 79.00 & 81.00 & 79.00& 98.00& 95.00& 68.00& 78.00& 88.00 & 95.00 & 84.00 & 80.00 & 90.00 & 91.00 & 92.00 & 99.00 & 95.00 & 97.00 & 90.00 & 87.00 & 98.00 \\ 
 & Part-only & 93.90 & 81.86 & 81.88 & 84.07& 98.72& 95.71& 70.63& 83.48& 87.97 & 95.16 & 95.83 & 87.53 & 91.73 & 95.05 & 95.92 & 99.46 & 97.19 & 97.93 & 90.26 & 86.20 & 96.65 \\
 uCelebA& Whole-only & 93.95 & 81.43 & 82.06 & 84.11& 98.57& 95.45& 70.66& 82.91& 89.08 & 95.52& 96.01 & 88.63 & 92.32 & 95.12 & 95.98 & 99.40 & 96.90 & 98.07 &90.67 & 86.57 & 97.10 \\
 & PaW & \textbf{94.64} &\textbf{ 83.01} & \textbf{82.86} & \textbf{84.58}& \textbf{98.93}& \textbf{95.93}& \textbf{71.46}& \textbf{83.63}& \textbf{89.84} & \textbf{95.85} & \textbf{96.11} & \textbf{88.50} & \textbf{92.62} & \textbf{95.46} & \textbf{96.26} & \textbf{99.59} & \textbf{97.38} & \textbf{98.21} & \textbf{91.53 }& \textbf{87.44} & \textbf{98.39} \\

 \hline\hline
& &\rot{Mouth Slightly Open} &\rot{Mustache} &\rot{Narrow Eyes} &\rot{No Beard} &\rot{Oval Face} &\rot{Pale Skin} &\rot{Pointy Nose} &\rot{Receding Hairline} &\rot{Rosy Cheeks} &\rot{Sideburns} &\rot{Smiling}&\rot{Straight Hair} &\rot{Wavy Hair} &\rot{Wearing Earrings} &\rot{Wearing Hat} &\rot{Wearing Lipstick} &\rot{Wearing Necklace} &\rot{Wearing Necktie} &\rot{Young} &\rot{}&\rot{\textbf{Average}} \\
 \hline

 &LNets+ANet~\cite{liu2015deep} & 92.00 & 95.00 & 81.00 & 95.00& 66.00& 91.00& 72.00& 89.00& 90.00 & 96.00 & 92.00 & 73.00 & 80.00 & 82.00 & 99.00 & 93.00 & 71.00 & 93.00 & 87.00 & &87.30 \\ 
 &Part-only & 93.55 & 96.63 & 86.96 & 95.71& 73.03 & 96.86& 76.40& 92.87& 94.77 & 97.63 & 91.98 & 82.53 & 81.29 & 89.07 & 98.75 & 92.96 & 87.13 & 96.69 & 86.51 &  & 90.46\\
uCelebA &Whole-only & 93.24 & 96.59 & 87.19 &95.40& 74.48 & 96.85& 76.06& 92.95& 94.83 & 97.50 & 91.61 & 82.18 & 82.63& 89.13 & 98.50 & 93.58 & 87.14 & 96.77 & 87.14 &  & 90.60\\
 &PaW & \textbf{94.05}& \textbf{96.90} & \textbf{87.56} & \textbf{96.22}& \textbf{75.03}& \textbf{97.08}& \textbf{77.35}& \textbf{93.44}& \textbf{95.07} & \textbf{97.64} & \textbf{92.73}& \textbf{83.52} & \textbf{84.07} & \textbf{89.93}& \textbf{99.02} & \textbf{94.24}& \textbf{87.70} & \textbf{96.85} & \textbf{88.59} &  & \textbf{91.23}\\
  
\hline 
\end{tabular}}
\label{tab:res}
\end{table*}

\subsubsection{PaW Classification Network}\label{sec:paw}
In this section, we evaluate the classification performance of the proposed PaW network.
Before showing the results, we first explore whether the RSL assigns appropriate weights to different subnets for attribute prediction and whether the ARL learns meaningful attributes correlations.

\noindent\textbf{Face Region Selection}
We visualize the weights of RSL in Figure~\ref{fig:comb_w}. 
Although each subnet predicts $M$ attribute scores simultaneously, only the weights of the corresponding part-based subnet against the whole-image-based subnet are shown here.
The weight magnitude indicates the importance of the subnet for predicting the attribute. 
Interestingly, we find that the part-based subnet related to the local attribute, \textit{e.g.} \textit{5 o Clock Shadow} and \textit{Bushy Eyebrows}, is always assigned the largest weight among the $M+1$ subnets. 
We also observe that for global attributes, \textit{e.g.} \textit{Attractive}, \textit{Blurry}, \textit{Heavy Makeup}, and \textit{Pale Skin}, the whole-image-based subnet achieves the highest weight. 
Intuitively those global attributes should obtain more information from the whole-image-based subnet. This validates the region selection ability of the RSL. 

\noindent\textbf{Face Attribute Correlation}
The learned ARL weights are visualized in Figure~\ref{fig:attr_corr}. 
We find that attribute pairs that are mutually exclusive such as (\textit{Attractive, Blurry}), (\textit{Black Hair, Blond Hair}) and (\textit{No Beard, Goatee}) are assigned lowest weights. Rarely co-occurring attribute pairs like (\textit{Male, Heavy Makeup}) are also assigned low weights. Pairs of attributes such as (\textit{Chubby, Double Chin}), (\textit{Heavy Makeup, Wearing Lipstick}) and (\textit{Smiling, High Cheekbones}) that commonly co-occur are given relatively higher weights. Moreover, the weights are asymmetric, for example, a person who wears lipstick is very unlikely to have a beard, but not the other way round. This is also reflected in the learned weights. This shows that ARL captures the attribute relationships.

\noindent\textbf{Classification Results}  
We show that our model achieves state-of-the-art results on uCelebA dataset.
In the following experiments, each subnet adopts the architecture of SNet3 in Table~\ref{tab:architect}.

We compare PaW with two baselines:

1. Part-only: each part net is trained on the detected face region to predict all face attributes. Then the attribute score from the most related part-based subnet is adopted for testing.

2. Whole-only: this method does not have part nets. It is trained with the whole face image only and is used to directly predict all attributes.

Table~\ref{tab:res} summarizes the classification performances. We observe that the PaW net performs consistently better than either the Part-only or Whole-only method alone, achieving an accuracy of 91.23\% vs. 90.60\% for Part-only and 90.46\% for Whole-only on uCelebA. This shows that RSL learns to selectively combine information from part-based and whole-image-based subnets. For unaligned face attribute classification on uCelebA dataset, we achieve the highest recognition rates across the board on all attributes and decrease the average recognition error from 12.70\% to 8.77\%, a reduction of 30.9\%. Our method on the aligned CelebA also achieves an accuracy of 91.33\% vs. 90.94\% compared with the state-of-the-art~\cite{rudd2016moon}.
This validates the effectiveness of the proposed attribute classification network. 
Also, the small performance gap on uCelebA and the aligned CelebA means that we practically eliminate the alignment step, and hence no special annotations are needed. 
Although the PaW network contains multiple part-based and whole-image-based subnets, the total number of parameters is only 11 M.

To test the importance of the FRL network, we further employ a baseline that divides each image into $4\times 4$ non-overlapping blocks to simulate crude part detectors. Then part-based subnets and whole-image-based subnet are trained the same way as before. It achieves an average accuracy of 90.95\% on uCelebA. However, we found that the weights corresponding to the whole-image-based net in the RSL are always higher than those of the part-based subnets for predicting \textit{all} the attributes. This is because coarse region localization makes the part-based subnets unreliable, thus all the predictions are essentially made by the whole-image-based subnet only. This validates the effectiveness of the proposed FRL network.

\section{Conclusions}
In this paper, we propose to learn 
attentional face regions to improve attribute classification performance under unaligned condition. To this end, a weakly-supervised face region localization network is first designed. Then the information from those detected regions are selectively combined by the hybrid classification network. Visualization shows our method not only discovers semantic meaningful attributes regions, but also captures rich correlations among attributes. Moreover, our results outperform state-of-the-art by a significant margin on the unaligned CelebA dataset. 

\bibliographystyle{aaai}
\bibliography{egbib}

\end{document}